\documentclass[runningheads]{llncs}
\usepackage{graphicx}
\usepackage{todonotes}
\usepackage{hyperref}
            
\usepackage{xcolor}

\usepackage{tikz}
\newcommand*\circled[1]{\tikz[baseline=(char.base)]{
            \node[shape=circle,draw,inner sep=2pt] (char) {#1};}}

\begin{document}
%
\title{\emph{New/s/leak} 2.0 -- Multilingual Information Extraction and Visualization for Investigative Journalism}
\titlerunning{Multilingual Info. Extraction and Visualization for Investigative Journalism}
%
\author{Gregor Wiedemann \and
Seid Muhie Yimam \and
Chris Biemann }
%
\authorrunning{G. Wiedemann et al.}
%
\institute{Language Technology Group\\ 
Department of Informatics, MIN Faculty\\ 
Universit\"{a}t Hamburg, Germany 
\email{\{gwiedemann,yimam,biemann\}@informatik.uni-hamburg.de}}

\maketitle              
\begin{abstract}
Investigative journalism in recent years is confronted with two major challenges: 1) vast amounts of unstructured data originating from large text collections such as leaks or answers to Freedom of Information requests, and 2) multi-lingual data due to intensified global cooperation and communication in politics, business and civil society. Faced with these challenges, journalists are increasingly cooperating in international networks. 
To support such collaborations, we present the new version of \emph{new/s/leak} 2.0, our open-source software for content-based searching of leaks. It includes three novel main features: 1) automatic language detection and language-dependent information extraction for 40 languages, 2) entity and keyword visualization for efficient exploration, and 3) decentral deployment for analysis of confidential data from various formats. We illustrate the new analysis capabilities with an exemplary case study.

\end{abstract}
\section{Investigative Journalism in the Digital Age}

In the era of digitization, journalists are confronted with major changes drastically challenging the ways how news are produced for a mass audience. Not only that digital publishing and direct audience feedback through online media influences what is reported on, how and by whom. But also digital data itself becomes a source and subject of newsworthy stories -- a development described by the term ``data journalism''. According to \cite{Simon.2017}, 51~\% of all news organizations in 2017 already employed dedicated data journalists and most of them reported a growing demand. The systematic analysis of digital social trace data blurs the line between journalism and (computational) social science. However, social scientists and journalists differ distinctively in their goals. Confronted with a huge haystack of digital data related to some social or political phenomenon, social scientists usually are interested in quantitatively or qualitatively characterizing this haystack while journalists actually look for the needle in it to tell a newsworthy story.
This `needle in the haystack' problem becomes especially vital for investigative stories confronted with large and heterogeneous datasets. 

Most of the information in such datasets is contained in written unstructured text form, for instance in scanned documents, letter correspondences, emails or protocols. Sources of such datasets typically range from 1) official disclosures of administrative and business documents, 2) court-ordered revelation of internal communication, 3) answers to requests based on Freedom of Information (FoI) acts, and 4) unofficial leaks of confidential information. In many cases, a public revelation of such confidential information benefits a democratic society since it contributes to reveal corruption and abuse of power, thus strengthening transparency of decisions in politics and the economy. 

To support this role of investigative journalism, we introduce the second, substantially re-engineered and improved version of our software tool \emph{new/s/leak} (``network of searchable leaks'') \cite{yimam-EtAl:2016:P16-4}. It is developed by experts from natural language processing and visualization in computer science in cooperation with journalists from ``Der Spiegel'', a large German news organization.\footnote{\url{https://www.spiegel.de}} \emph{New/s/leak} 2.0 serves three central requirements that have not been addressed sufficiently by previously existing solutions for investigative and forensic text analysis:
\begin{enumerate}
    \item Since journalists are primarily interested in stories around persons, organizations and locations, we adopt a visual exploration approach centered around named entities and keywords.
    \item Many tools only work for English documents or a small number of other `big languages'. To foster international collaboration, our tool allows for simultaneous analysis of documents from a set of currently 40 languages.
    \item The work with confidential data such as from unofficial leaks requires a decentralized analysis environment, which can be used potentially disconnected from the internet. We distribute \emph{new/s/leak} as a free, open-source server infrastructure via Docker containers, which can be easily deployed by both news organizations and single journalists. 
\end{enumerate}
In the following sections, we introduce related work and discuss technical aspects of \emph{new/s/leak}. We also illustrate its analysis capabilities in a brief case study.


\section{Related Work}
\label{related}
A number of commercial and open-source software products have been developed to support data journalists in their work with large datasets. 
Many tools such as OpenRefine, Datawrapper, or Tabula concentrate on structured data only. For unstructured text data, there are only few options. Since licenses for commercial tools often are prohibitively expensive especially for smaller news agencies or individuals, we focus on open-source products in the following comparison.

The most reputated application is \textit{DocumentCloud}\footnote{\url{https://www.documentcloud.org}}, an open platform designed for journalists to annotate and search in (scanned) text documents. It is supported by popular media partners such as the New York Times, and PBS. Besides fulltext search it provides automatic named entity recognition (NER) based on OpenCalais \cite{ThomsonReuters.2017} in English, Spanish and German. 

\textit{Overview} \cite{Brehmer2014} is a more advanced open-source application developed by computer scientists in collaboration with journalists to support investigative journalism. The application supports import of PDF, MS Office and HTML documents, document clustering based on topic similarity, a simplistic location entity detection, full-text search, and document tagging. Since this tool is already mature and has successfully been used in a number of published news stories, we adapted some of its most useful features such as document tagging and a keyword-in-context (KWIC) view for search hits. Furthermore, in \emph{new/s/leak} we concentrate on intuitive and visually pleasing approaches to display extracted contents for a fast exploration of collections. 

The \textit{Jigsaw visual analytics} \cite{Stasko:2008:JSI:1466620.1466622,Gorg.2013,Carsten2014} system is a third tool that supports investigative analysis of textual documents. Jigsaw focuses on the extraction of entities (using multiple NER tools for English) and their correlation with metadata. It provides visualization of entities as lists and document contents as a (tree-structured) word graph. \emph{New/s/leak} instead visualizes coherence of such information as network graphs.

Recent investigations on leaked data from off-shore money laundering companies, usually referred to as ``Panama Papers'' \cite{ODonovan.2016} and ``Paradise Papers'' \cite{Obermayer.2017} lead to the development of two software packages, which was driven by exactly those uses cases: \textit{Aleph}\footnote{\url{https://github.com/alephdata/aleph}} and \textit{DataShare}\footnote{\url{https://github.com/ICIJ/datashare}}. Both packages are designed to support typical  'follow-the-money' investigations to reveal corruption. Hence, they put emphasis on integrating information extraction from text such as named entities with structured databases containing additional information about entities. 

\emph{New/s/leak} differs from the aforementioned software products mainly with respect to its focus on visual exploration of unknown and unstructured datasets. The idea of visualizing entity networks to explore large text collections for investigative journalism was already realized in the first version of our tool \cite{yimam-EtAl:2016:P16-4}. But, the software had several limitations such as 1) it only allowed processing of monolingual datasets in either English or German, 2) it only supported a predefined number of entity and meta-data types, and 3) it did not support entity extraction with user-defined dictionaries and rule patterns. Moreover, it lacked a user-friendly data import process leaving the burden of data wrangling of heterogeneous document collections to the journalist.

\section{Architecture}
\label{arch}
\begin{figure}[t]
    \centering	\includegraphics[width=0.9\linewidth,trim={0 3.3cm 3cm 0},clip]{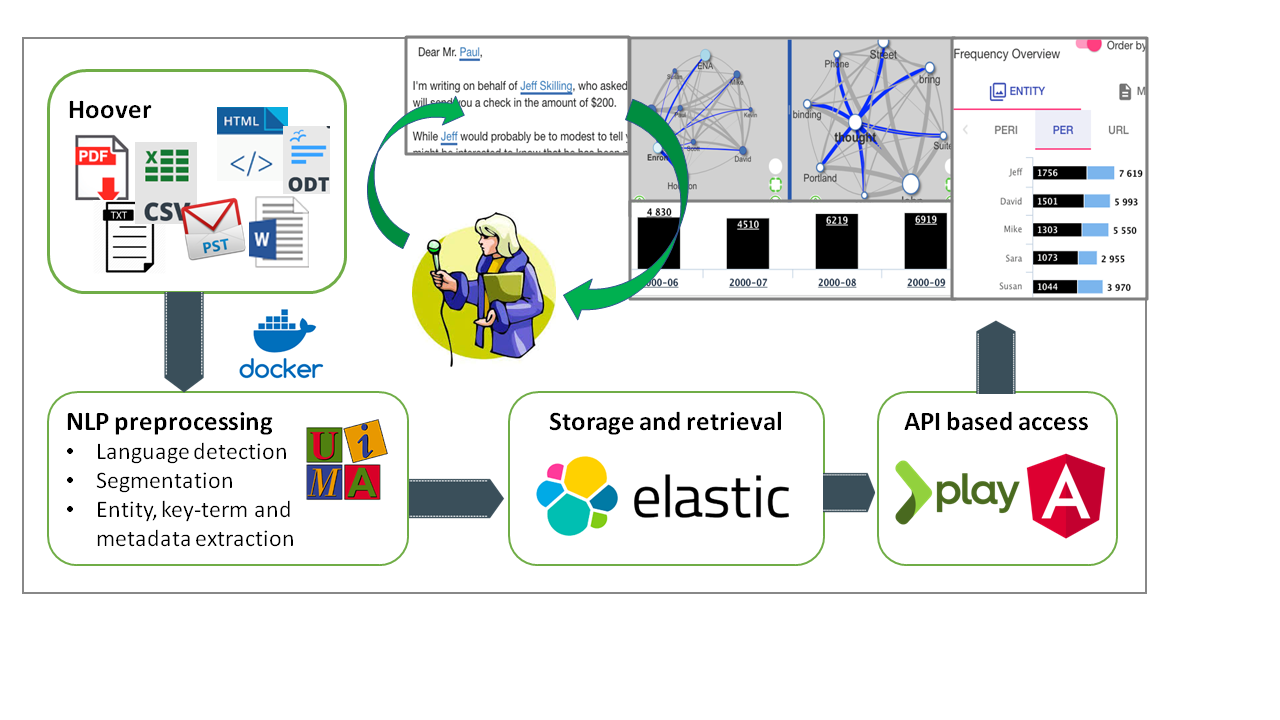}
	\caption{``Dockerized'' architecture of \emph{new/s/leak} 2.0}
	\label{fig:arch} 
	\end{figure}

Figure \ref{fig:arch} shows the architecture of \emph{new/s/leak}. In order to allow users to analyze a wide range of document types, our system includes a document processing component, which extracts text and metadata from a variety of document types into a unified index representation. 

On the extracted texts, a number of NLP processing tasks are performed in a UIMA pipeline \cite{Ferrucci.2004b}: automatic identification of the document language, segmentation of documents into appropriate paragraph, sentence and token units, and extraction of named entities, keywords and metadata. Fulltexts and extracted data are stored in an ElasticSearch index\footnote{\url{https://www.elastic.co}}. Access and real-time analysis of the data is provided by RESTful web services based on the Scala Play framework. We offer interactive information visualization technologies based on the D3 visualization library \cite{Bostock.2011} composed into an AngularJS browser application. 

In order to enable a seamless deployment of the tool for journalists with limited technical knowledge, we integrated all of the required components of the architecture into a Docker\footnote{\url{https://www.docker.com}} setup. Via ``docker-compose'', a software to orchestrate Docker containers for complex architectures, end-users can download and run locally a preconfigured version of \emph{new/s/leak} with only a few commands. Relying on Docker also ensures, that it runs on any Docker-compatible operating system including Linux, macOS, and Windows. Being able to process data locally and even without any internet connection is a vital prerequisite for journalists, especially if they work with sensitive leaks or FoI data. All necessary source code and installation instructions can be found on our Github page.\footnote{\url{https://uhh-lt.github.io/newsleak}}

\section{Data Wrangling}
\label{sec:data_wrangling}

Extracting text and metadata from various formats into a format readable by a specific analysis tool can be a tedious task. In an investigative journalism scenario it can even be a deal breaker since time is an especially valuable resource and file format conversion might not be a task journalists are well trained in. 
To offer easy access to as many file formats as possible in \emph{new/s/leak}, we opted for a close integration with \emph{Hoover},\footnote{\url{https://hoover.github.io}} a set of open-source tools for text extraction and search in large text collections. Hoover is developed by the European Investigative Collaborations (EIC) network\footnote{\url{https://eic.network}} with a special focus on large data leaks and heterogeneous data sets. It can extract data from various text file formats (txt, html, docx, pdf) but also extracts from archives (zip, tar, etc.) and email inbox formats (pst, mbox, eml). The text is extracted along with metadata from files (e.g. file name, creation date, file hash) and header information (e.g. subject, sender, receiver in case of emails). Extracted data is stored in an ElasticSearch index. Then, \emph{new/s/leak} connects directly to Hoover's index to read full texts and metadata for its own information extraction pipeline.

\section{Multilingual Information Extraction}
\label{sec:multilang_ner}

The core functionality of \emph{new/s/leak} to support investigative journalists is the automatic extraction of various kinds of entities from text to facilitate the exploration and the sense-making process from large collections. Since a lot of steps in this process involve language dependent resources, we put emphasis on supporting as many languages as possible.

\noindent{\textbf{Preprocessing:}}
Information extraction is implemented as a configurable UIMA pipeline \cite{Ferrucci.2004b}. Text documents and metadata from a Hoover collection (see Section~\ref{sec:data_wrangling}) are read in parallelized manner and put through a chain of automatic annotators. In the final step of the chain, results from annotation processes are indexed in an ElasticSearch index for later retrieval and visualization.
First, we identify the language of each document. 
Second, we separate sentences and tokens in each text. To guarantee compatibility with various Unicode scripts in different languages, we rely on the ICU4J library\cite{icu4j.2018}, which provides sentence and word boundary detection. 

\noindent{\textbf{Dictionary and pattern matching:}}
In many cases, journalists follow some hypothesis to test for their investigative work. Such a proceeding can involve looking for mentions of already known terms or specific entities in the data. This can be realized by lists of dictionaries provided to the initial information extraction process. \emph{New/s/leak} annotates every mention of a dictionary term with its respective list type. Dictionaries can be defined in a language-specific fashion, but also applied across all languages in the corpus. Extracted dictionary entities are visualized later on along with extracted named entities.
In addition to self-defined dictionaries, we annotate email addresses, telephone numbers, and URLs with regular expression patterns. This is useful, especially for email leaks, to reveal communication networks of persons.

\noindent{\textbf{Temporal expressions:}}
Tracking documents across time of their creation can provide valuable information during investigative research.
Unfortunately, many document sets (e.g. collections of scanned pages) do not come with a specific document creation date as structured metadata. To offer a temporal selection of contents to the user, we extract mentions of temporal expressions in documents. This is done by integrating the Heideltime temporal tagger \cite{Strotgen.2015} in our UIMA workflow. Heideltime provides automatically learned rules for temporal tagging in more than 200 languages. Extracted timestamps can be used to select and filter documents.

\noindent{\textbf{Named Entity Recognition:}}
We automatically extract person, organization and location names from all documents to allow for an entity-centric exploration of the data collection. Named entity recognition is done using the \emph{polyglot-NER} library \cite{AlRfou.2015}. Polyglot-NER contains sequence classification for named entities based on weakly annotated training data automatically composed from Wikipedia\footnote{\url{https://wikipedia.org}} and Freebase\footnote{\url{https://developers.google.com/freebase}}. Relying on the automatic composition of training data allows polyglot-NER to provide pre-trained models for 40 languages (see Appendix).


\noindent{\textbf{Keyterm extraction:}}
To further summarize document contents in addition to named entities, we automatically extract keyterms and phrases from documents. For this, we have implemented our own keyterm extraction library. The approach is based on statistical comparison of document contents with generic reference data. Reference data for each language is retrieved from the Leipzig Corpora Collection \cite{Goldhahn.2012}, which provides large representative corpora for language statistics. We included resources for the 40 languages also covered by the NER library. We employ log-likelihood significance as described in \cite{Rayson.2004} to measure the overuse of terms (i.e. keyterms) in our target documents compared to the generic reference data. Ongoing sequences of keyterms in target documents are concatenated to key phrases if they occur regularly in that exact same order. Regularity is determined with the Dice statistic, which allows reliably to extract multiword units such as ``stock market'' or ``machine learning'' in the documents. 
Since the keyterm extraction method may also extract named entities there can be a substantial overlap between the two types. To allow for a separate display of entities and keywords in a later step, we ignore keyterms that already have been identified as named entities.
The remaining top keyterms are used to provide a brief summary of each document for the user.
    
\begin{figure}[t]
    \centering
	\includegraphics[width=0.8\linewidth]{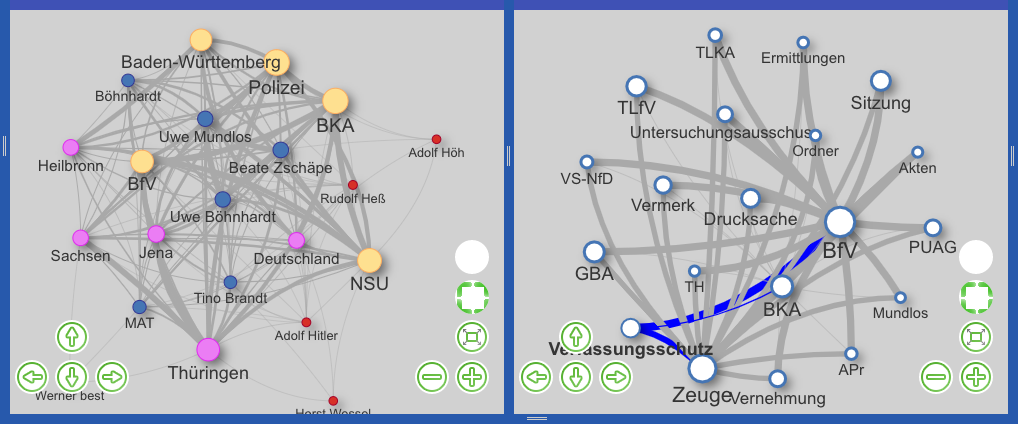}
	\caption{The entity and keyword graphs based on the parliamentary reports (see Section~\ref{casestudy}). Networks are visualized interactively based on the current document selection, which can be filtered by fulltext search, entities or metadata. 
	}
	\label{fig:entkey} 
	\end{figure}

\noindent{\textbf{Entity- and Keyword-Centric Visualization and Filtering:}}
Access to unstructured text collections via named entities and keyterms is essential for journalistic investigations. To support this, we included two types of graph visualization, as it is shown in Figure \ref{fig:entkey}. The first graph, called entity network, displays entities in a current document selection as nodes and their joint occurrence as edges between nodes. Different node colors represent different types such as person, organization or location names. Furthermore, mentions of entities that are annotated based on dictionary lists or annotated by a given regular expression are included in the entity network graph. The second graph, called keyword network, is build based on the set of keywords representing the current document selection. 

Besides to fulltext search, visualizations are the core concept to navigate in an unknown dataset. Nodes in networks as well as entities, metadata or document date ranges displayed as frequency bar charts in the user interface (see Appendix) can be added as filter to constrain the current set of documents. By this, journalists can easily drill down into interesting aspects of the dataset or zoom out again by removing filter conditions. From current sub-selections, user can easily switch to the fulltext view of a document, which also highlights extracted entities and allows for tagging of interesting parts.

\section{Exemplary Case Study: Parliamentary Investigations}
\label{casestudy}
In the following, we present an exemplary case studies to illustrate the analysis capabilities of our tool.
The scenario is centered around the parliamentary investigations of the ``Nationalsozialistischer Untergrund'' (NSU) case in Germany. Between 1998 and 2011, a terror network of neo-Nazis was responsible for murders of nine migrants and one policewoman. Although many informants of the police and the domestic intelligence service (Verfassungsschutz) were part of the larger network, the authorities officially did neither take notice of the group nor the racist motives behind the murder cases. Since 2012, this failure of domestic intelligence led to a number of parliamentary investigations. We collected 7 final reports from the Bundestag and federal parliaments investigating details of the NSU case. Altogether the reports comprise roughly 12,000 pages.

In \emph{new/s/leak}, long documents such as the reports can be split into more manageable units of certain length such as paragraphs or pages. Smaller units ensure that co-occurrence of extracted entities and keywords actually represents semantic coherence. By splitting into sections of an average page size as a minimum, we receive 12,021 analysis units. For our investigation scenario, we want to follow a certain hypothesis in the data. Since the NSU core group acted against a background of national socialist ideology, we try to answer the following question: To which extent did members of the neo-Nazi network associate themselves with protagonists of former Nazi Germany?

To answer this question, we feed a list of prominent NSDAP party members collected from Wikipedia as additional dictionary into the information extraction process. The resulting entity network reveals mentioning of 17 former Nazis in the reports with Rudolf He{\ss} as the most frequent one. Heß is celebrated as a martyr in the German neo-Nazi scene. In the third position, the list reveals Adolf H{\"o}h, a much less prominent NSDAP party member. Filtering down the collection to documents containing reference to this person reveals the context of the NSU murder case in Dortmund 2006. At that time an openly acting neo-Nazi group existed in Dortmund, which named itself after the former SA-member H{\"o}h who got killed by communists in 1930. The network also reveals connections to another SA-member, Walter Spangenberg, who got killed in Cologne during the 1930s as well. The fact that places of the two killings, Dortmund and Cologne, are near places of two later NSU attacks led to a specific theory about the patterns of the NSU murders in the parliamentary investigation. As the keyterm network reveals, the cases are discussed with reference to the term ``Blutzeugentheorie'' ('lit. trans: theory of blood witnesses') in the reports. Our combination of an external name list with the automatic extraction of location entities and keyterms quickly led us to this striking detail of the entire case.

\section{Discussion}

In this article, we introduced version 2.0 of \emph{new/s/leak}, a software to support investigative journalism. 
It aims to support journalists throughout the entire process of analyzing large, unstructured and heterogeneous document collections: data cleaning and formatting, metadata extraction, information extraction, interactive filtering, close reading, an tagging interesting findings. 
We reported on technical aspects of the software based on the central idea to approach an unknown collection via extraction and display of entity and keyterm networks. As a unique feature, our tool allows a simultaneous analysis of multi-lingual collections for up to 40 languages.
We further demonstrated in an exemplary use case that the software is able to support investigative journalists in the two major ways of proceeding research (cp. \cite{Brehmer2014}): 1. exploration and story finding in an unknown collection, and 2. testing hypothesis based on previous information.
Cases such as ``Paradise Papers'', ``Dieselgate'' or ``Football leaks'' made it especially clear that a decentralized collaboration tool for analysis across different languages is needed for effective journalistic investigation. We are convinced that our software will contribute to this goal in the work of journalists at the ``Der Spiegel'' news organization and, since it is released as open-source, also for other news organizations and individual journalists.

\paragraph{Acknowledgement} The work was funded by Volkswagen Foundation under Grant Nr. 90 847. 

\bibliographystyle{splncs04}
\bibliography{multinewsleak}

\newpage
\section*{Appendix}

\begin{figure}
    \centering
	\includegraphics[width=0.99\linewidth,height=20em]{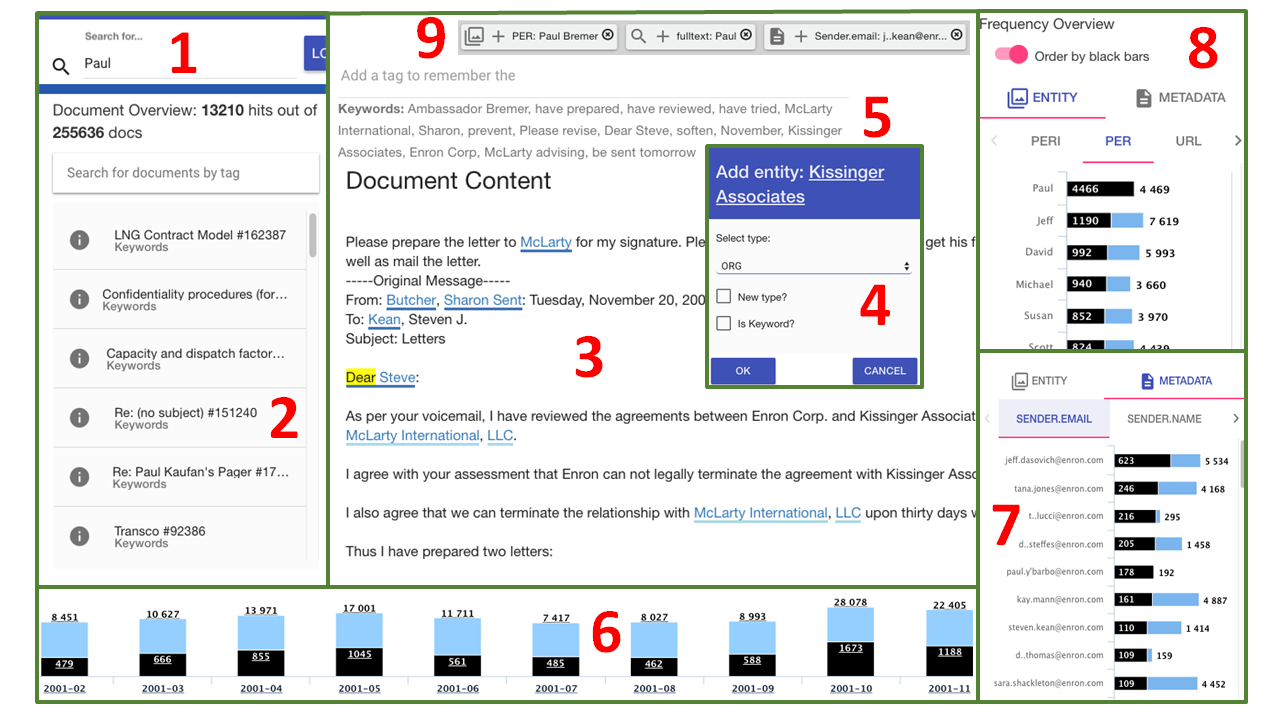}
	\caption{Different UI components of the \emph{new/s/leak} 2.0 system.}
	\label{fig:docview} 
\end{figure}

\paragraph{\textbf{New/s/leak UI Components:}} UI components are designed to help journalists discovering interesting stories more easily. Figure \ref{fig:docview} presents different UI components that facilitate the close reading. The numbers in Figure \ref{fig:docview} mark the different types of visual components supported in our tool. \circled{{\color{red}1}} shows the full-text searching component, \circled{{\color{red}2}} presents the list of documents that are retrieved based on the current set of filters. In \circled{{\color{red}3}}, the user can read the actual document content, but also annotate new entities (\circled{{\color{red}4}}), which were not detected by the NER, merging different textual references to the same entities, and remove/blacklist wrongly detected entities. The possibility to tag individual documents can be used to organize the collection with respect to any user-defined category system. \circled{{\color{red}5}} shows lists of keywords summarizing the document. \circled{{\color{red}6}}, \circled{{\color{red}7}}, and \circled{{\color{red}8}} displays temporal evolution, metadata and entity charts to help the reader to further filter documents. Lastly, \circled{{\color{red}9}} shows a history of search filters applied to the document collection so far. Not shown here: graph visualization of current document set, as depicted in Figure \ref{fig:entkey}. 

\paragraph{\textbf{Supported Languages:}} Arabic, Bulgarian, Catalan, Chinese, Croatian, Czech, Danish, Dutch, English, Estonian, Finnish, French, German, Greek, Hebrew, Hindi, Hungarian, Indonesian, Italian, Japanese, Korean, Latvian, Lithuanian, Malay, Norwegian, Persian, Polish, Portuguese, Romanian, Russian, Serbian, Slovak, Slovene, Spanish, Swedish, Tagalog, Thai, Turkish, Ukrainian, Vietnamese

\end{document}